\documentclass[runningheads]{llncs}

\usepackage[T1]{fontenc}
\usepackage{graphicx}
\usepackage{bbding}
\usepackage{cite}
\usepackage{amsmath,amssymb,amsfonts}
\usepackage{algorithm}
\usepackage[noend]{algpseudocode}
\usepackage{graphicx}
\usepackage{textcomp}
\usepackage{xcolor}
\usepackage{xspace}
\usepackage[labelfont=scriptsize,textfont=scriptsize]{subfig}

\newcommand{\sysname}{\textsf{\textsc{SARMTO}}\xspace}

\begin{document}
\title{ Secure Resource Allocation via Constrained Deep Reinforcement Learning }

\author{
Jianfei Sun \inst{1}
\and 
Qiang Gao  \inst{2} 
\and
Cong Wu \inst{3}\Envelope
\and
Yuxian Li \inst{1} 
\and \\
Jiacheng Wang \inst{3}
\and 
Dusit Niyato\inst{3}
}
\institute{ \scriptsize
School of Computing \& Information Systems, Singapore Management University, 188065, Singapore
\and
School of Computing \& Artificial Intelligence, Southwestern University of Finance and Economics, Chengdu 611130, China\\
\and
College of Computing \& Data Science, Nanyang Technological University, 639798, Singapore\\
\email{\{jfsun,yuxianli\}@smu.edu.sg, qianggao@swufe.edu.cn, and \\ \{cong.wu,jiacheng.wang, dniyato\}@ntu.edu.sg\\} 
}
\maketitle      
\thispagestyle{empty}
\pagestyle{empty}

\begin{abstract}

The proliferation of Internet of Things (IoT) devices and the advent of 6G technologies have introduced computationally intensive tasks that often surpass the processing capabilities of user devices. Efficient and secure resource allocation in serverless multi-cloud edge computing environments is essential for supporting these demands and advancing distributed computing. However, existing solutions frequently struggle with the complexity of multi-cloud infrastructures, robust security integration, and effective application of traditional deep reinforcement learning (DRL) techniques under system constraints. To address these challenges, we present \sysname, a novel framework that integrates an action-constrained DRL model. \sysname dynamically balances resource allocation, task offloading, security, and performance by utilizing a Markov decision process formulation, an adaptive security mechanism, and sophisticated optimization techniques. Extensive simulations across varying scenarios—including different task loads, data sizes, and MEC capacities—show that \sysname consistently outperforms five baseline approaches, achieving up to a 40\% reduction in system costs and a 41.5\% improvement in energy efficiency over state-of-the-art methods. These enhancements highlight \sysname's potential to revolutionize resource management in intricate distributed computing environments, opening the door to more efficient and secure IoT and edge computing applications.

\end{abstract}

\section{Introduction}

The rapid expansion of Internet of Things (IoT) devices and the advent of 6G technologies~\cite{lin2024efficient,yuan2025constructing,lin2025leo,liang2024vulseye,yuan2023graph} have led to a surge in computationally demanding applications, such as augmented reality~\cite{he2020optimizing,estrada2022deep,li2022integrated}, autonomous vehicles~\cite{lin2022tracking,fang2024ic3m,lin2022channel}, and real-time data analytics~\cite{wu2024rethinking}, which often exceed the processing capabilities of user devices~\cite{zhang2024fedac}. To address this, serverless multi-cloud edge computing has emerged as a promising solution, offering flexible, scalable, and proximity-aware resources~\cite{wang2022multi,zhang2024secure,lin2024hierarchical,wang2024adaptive}. However, the heterogeneous and distributed nature of these environments as well as user mobility pose significant challenges to secure and efficient resource allocation~\cite{macfl2022,liang2024resource,lin2024fedsn,zhang2023privacyeafl}.

Significant progress has been made in addressing various aspects of resource allocation challenges. Tang et al.~\cite{tang2022distributed} explored distributed task scheduling using dueling double deep Q-networks for optimizing resource allocation in edge computing. Yao et al.~\cite{yao2023performance} developed an experience-sharing deep reinforcement learning (DRL) method for function offloading in serverless edge environments. In terms of security, Min et al.~\cite{min2018learning} introduced a privacy-aware offloading scheme using reinforcement learning for healthcare IoT devices, while Huang et al.~\cite{huang2019security} proposed a secure and energy-efficient offloading strategy for service workflows in mobile edge computing. Additionally, Ko et al.~\cite{ko2021performance} and Cicconetti et al.~\cite{cicconetti2020architecture} investigated the integration of serverless computing with edge environments to enhance scalability and resource utilization.

Despite these advancements, critical research gaps persist. First, most solutions target single-cloud edge models, overlooking the complexities of multi-cloud environments with heterogeneous nodes~\cite{grozev2014multi,fang2024automated}. Second, while DRL shows promise in resource management, current methods often rely on penalty-based rewards, resulting in suboptimal policies and convergence issues~\cite{zhang2024security}. Third, the integration of robust security with efficient resource allocation in serverless multi-cloud environments is largely unexplored, particularly in the context of dynamic threats and varying task sensitivities~\cite{elgendy2020efficient}. These gaps underscore the need for a comprehensive framework that securely and efficiently allocates resources in complex, heterogeneous serverless multi-cloud edge environments.

In this paper, we propose secure adaptive resource management and task optimization (\sysname), a novel framework for secure and efficient resource allocation in serverless multi-cloud edge environments. At its core, \sysname employs an innovative action-constrained deep reinforcement learning (AC-DRL) model that optimizes task offloading and resource allocation while respecting system constraints. Using a Markov decision process (MDP) formulation, it models the resource allocation problem as a sequential decision-making process, allowing to adapt to changing conditions and balance objectives such as latency, energy efficiency, and security.

\sysname addresses key challenges through several novel design elements. First, it introduces a flexible state representation and action space to manage the heterogeneity of computing nodes and the complexity of multi-cloud environments. Second, an action constraint mechanism enforces system limitations during decision-making, avoiding the drawbacks of penalty-based approaches. 
Third, an adaptive security mechanism dynamically adjusts protection levels based on task sensitivity and threat landscape, ensuring robust security without compromising efficiency. Finally, advanced techniques including prioritized experience replay and dueling network architecture are integrated to enhance learning efficiency in this complex decision space.

In summary, our contributions are as follows:
\begin{itemize}
 \item We propose \sysname, a novel framework for secure and efficient resource allocation in serverless multi-cloud edge environments.

\item We introduce AC-DQN, a novel action-constrained DRL algorithm that respects system constraints and dynamically balances security with computational overhead.

\item We evaluate \sysname through extensive simulations, showing significant improvements in performance, energy efficiency, and security over existing methods.
\end{itemize}

\section{Related Work}

Recent advances in edge-cloud computing have led to significant progress in optimizing resource allocation. Tang et al.~\cite{tang2022distributed} developed a distributed task scheduling algorithm using dueling double deep Q-networks for heterogeneous edge networks, while Yao et al.~\cite{yao2023performance} introduced a DRL-based function offloading method for serverless edge computing, achieving better latency and success rates. However, both approaches fall short in addressing critical security concerns and managing energy costs effectively.

DRL has emerged as a powerful tool for dynamic resource management in complex environments. Xu et al.~\cite{xu2020service} applied DRL with deep Q-networks to optimize service offloading in vehicular edge computing, and Chen et al.~\cite{chen2021multiuser} extended this by improving resource allocation in cloud-edge networks. Despite these advancements, many DRL-based methods rely on penalty functions to enforce constraints, which can lead to suboptimal outcomes. To address this, our work introduces an action constraint mechanism that integrates constraints directly during decision-making. Furthermore, while serverless computing in edge environments has been explored for its scalability and efficiency, as demonstrated by Ko et al.~\cite{ko2021performance} and Cicconetti et al.~\cite{cicconetti2020architecture}, the security challenges and complexities of multi-cloud scenarios remain underexplored. Our research focuses on bridging these gaps by developing a secure resource allocation framework for serverless multi-cloud edge environments, accounting for heterogeneous nodes and dynamic

\section{System Model and Problem Formulation}
\begin{figure}[!t]
	\centering
	\includegraphics[width = 0.95\linewidth]{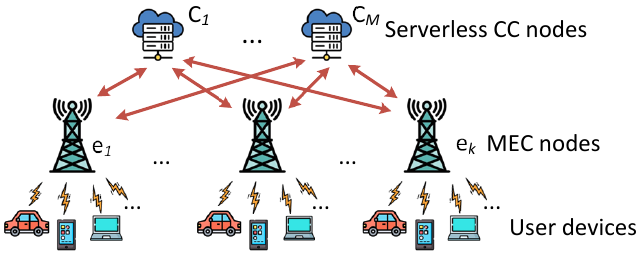}
	\caption{Serverless multi-cloud edge computing model}
	\label{fig:sysmodel}
\end{figure}
As illustrated in Figure~\ref{fig:sysmodel}, we consider a heterogeneous serverless multi-cloud edge computing network $\Gamma = (U, E, C)$. The network comprises a set of user devices (UDs) $U = \{u_1, ..., u_N\}$, multi-access edge computing (MEC) nodes $E = \{e_1, ..., e_K\}$, and cloud computing (CC) nodes $C = \{c_1, ..., c_M\}$. The network topology is represented by a weighted graph $G(V, E)$, where $V = U \cup E \cup C$ and $E$ represents the set of communication links. Each link $(v_1, v_2) \in E$ is characterized by its bandwidth $B_{v_1v_2}$ and channel gain $g_{v_1v_2}$.

\textbf{Task model:}
Each UD $u_i \in U$ generates a set of computational tasks $T_i = \{\tau_{i1}, ..., \tau_{iL}\}$. A task $\tau_{ij}$ is defined by the tuple $(D_{ij}, C_{ij}, T_{ij})$, where $D_{ij} \in \mathbb{R}^+$ denotes the input data size in bits, $C_{ij} \in \mathbb{Z}^+$ represents the required CPU cycles for computation, and $T_{ij} \in \mathbb{R}^+$ is the delay constraint in seconds. The relationship between $C_{ij}$ and $D_{ij}$ is modeled as:
\begin{equation}
	C_{ij} = \zeta(D_{ij}),
\end{equation}
where $\zeta(\cdot)$ is a function that maps data size to required CPU cycles, based on the application type as shown in Table~\ref{tab:complexity}~\cite{zhang2024security}.

\begin{table}[h]
	\centering
	\caption{Application complexity.}
	\begin{tabular}{|l|c|c|}
		\hline
		\textbf{Application}             & \textbf{Labels} & \textbf{CPU cycle/Byte} \\ \hline
		Gzip                              & A               & 330                     \\ \hline
		Health monitoring                 & B               & 500                     \\ \hline
		pdf2text (N900 data sheet)        & C               & 960                     \\ \hline
		x264 CBR encode                   & D               & 1900                    \\ \hline
		html2text                         & E               & 5900                    \\ \hline
		Pdf2text (E72 data sheet)         & F               & 8900                    \\ \hline
		Augmented reality                 & G               & 12000                   \\ \hline
	\end{tabular}
	\label{tab:complexity} 
\end{table}

\textbf{Computation resources:}
Computation nodes include MEC nodes and CC nodes. Each MEC node $e_k \in E$ is characterized by its computational capacity $f_m^k$ (in CPU cycles/second) and unit energy cost $\mu_m^k$. Similarly, each CC node $c_j \in C$ is defined by its computational capacity $f_c^j$ and unit energy cost $\mu_c^j$. These parameters capture the heterogeneity of the computing resources available in the network.

Our objective is to find an optimal task offloading and resource allocation policy $\pi^*$ that minimizes the overall system cost while satisfying security and delay constraints. We introduce a binary decision variable $x_{ijk} \in \{0, 1\}$ indicating whether task $\tau_{ij}$ is executed on node $k$ (either MEC or CC). The optimization problem is formulated as:
\begin{equation}
	\begin{aligned}
		\min_{\pi} \quad & C(\pi) = \alpha_1 T(\pi) + \alpha_2 E(\pi) \\
		\text{s.t.:} \quad & \sum_{k \in E \cup C} x_{ijk} = 1, \quad \forall i, j \\
		& T_{ij}(\pi) \leq T_{ij}, \quad \forall i, j \\
		& \sum_{i,j} x_{ijk}C_{ij} \leq f_k, \quad \forall k \in E \cup C \\
		& x_{ijk} \in \{0, 1\}, \quad \forall i, j, k
	\end{aligned}
\end{equation}
where $C(\pi)$ represents the total system cost under policy $\pi$, which is a weighted sum of the total delay $T(\pi)$ and total energy consumption $E(\pi)$. The weights $\alpha_1$ and $\alpha_2$ ($\alpha_1 + \alpha_2 = 1$) allow for flexible prioritization between delay and energy objectives. $T_{ij}(\pi)$ denotes the actual execution time of task $\tau_{ij}$ under policy $\pi$, and $f_k$ is the computational capacity of node $k$.

\textbf{Delay and energy model:}
The total delay $T(\pi)$ is calculated as the sum of computation time, communication time (if offloaded), and security overhead for all tasks, which is given as:
\begin{equation}
T(\pi) = \sum_{i,j,k} x_{ijk}(T_{ij}^{\text{comp}} + T_{ij}^{\text{comm}} + T_{ij}^{\text{sec}}).
\end{equation}

Similarly, the total energy consumption $E(\pi)$ is given by:
\begin{equation}
E(\pi) = \sum_{i,j,k} x_{ijk}(E_{ij}^{\text{comp}} + E_{ij}^{\text{comm}} + E_{ij}^{\text{sec}}),
\end{equation}
where $E_{ij}^{\text{comp}}$, $E_{ij}^{\text{comm}}$, and $E_{ij}^{\text{sec}}$ represent computation energy, communication energy (if offloaded), and security-related energy consumption, respectively.

\textbf{Security overhead model:}
The security overhead, i.e., due to cryptographic methods, is an essential aspect of our model, capturing the additional time and energy costs associated with ensuring data integrity and privacy, which are modeled as:
\begin{equation}
T_{ij}^{\text{sec}} = (\phi_{ij}^{\text{enc}} + \omega_{ij}) / f_k + (\phi_{ij}^{\text{dec}} + \omega_{ij}) / f_{k'},
\end{equation}
\begin{equation}
E_{ij}^{\text{sec}} = \mu_k(\phi_{ij}^{\text{enc}} + \omega_{ij}) + \mu_{k'}(\phi_{ij}^{\text{dec}} + \omega_{ij}),
\end{equation}
where $\phi_{ij}^{\text{enc}}$, $\phi_{ij}^{\text{dec}}$, and $\omega_{ij}$ represent the CPU cycles required for encryption, decryption, and integrity verification, respectively. The nodes $k$ and $k'$ denote the source and destination nodes for offloading.


\section{Design of \sysname}
\subsection{Overview}
\sysname is a cutting-edge framework designed to optimize secure resource allocation in serverless multi-cloud edge environments.It employs an action-constrained DRL model that integrates five components: an MDP model for sequential decision-making, an action-constrained deep Q-network (AC-DQN) to respect system constraints, a robust security mechanism for data protection, an adaptive exploration strategy for efficient policy learning, and advanced performance optimization techniques. This enables to dynamically adapt to heterogeneous computing conditions, effectively balancing task offloading, resource allocation, security, and performance requirements in complex distributed computing systems.

\subsection{MDP Formulation}

The resource allocation problem in serverless multi-cloud edge environments is inherently dynamic and sequential. By modeling it as an MDP, we can capture the temporal dependencies and uncertainties inherent in the system, allowing for more effective decision-making over time. This approach enables \sysname to consider the long-term impact of allocation decisions rather than just immediate rewards, leading to more robust and efficient resource management.

MDP is defined as a tuple $M = \{S, A, P, R, \gamma\}$, where $S$ represents the state space, $A$ the action space, $P$ the state transition probabilities, $R$ the reward function, and $\gamma$ the discount factor. The state $s_t \in S$ at time $t$ is defined as $s_t = \{D_{i,t}, C_{i,t}, T_{i,t} \,|\, i = 1, \ldots, N\}$, capturing the data size, CPU cycles, and delay constraints for all tasks, thereby allowing the model to account for all relevant factors in allocation decisions. The action $a_t \in A$ is represented as $a_t = \{x_{i,k,t} \,|\, i = 1, \ldots, N; k \in E \cup C\}$, specifying task allocation decisions to either edge or cloud nodes. The reward $r_t$ is defined as the negative of the system cost: $r_t = -C_t = -(\alpha_1 T_t + \alpha_2 E_t)$, enabling \sysname to optimize both performance and energy efficiency. The weights $\alpha_1$ and $\alpha_2$ provide the flexibility to adjust the trade-off between these objectives.

Algorithm~\ref{alg:mdp_formulation} outlines the MDP formulation process for \sysname. It begins by defining the MDP tuple and initializing its components: the state space, action space, transition probabilities, reward function, and discount factor. For each time step, the algorithm observes the current state, which includes the data size, CPU cycles, and delay constraints of all tasks. Based on this state, an action is chosen, representing the task allocation decisions for each task to either edge or cloud nodes. After executing the action, the algorithm observes the new state and calculates the reward, which is defined as the negative of the system cost, balancing delay and energy consumption. Finally, the MDP model is updated based on the observed transition and reward, allowing \sysname to refine its decision-making process over time. This iterative process enables \sysname to adapt to the dynamic nature of the serverless multi-cloud edge environment and continuously improve its resource allocation strategy.
\begin{algorithm}[!t]
	\caption{MDP Formulation for \sysname}
	\label{alg:mdp_formulation}
	\begin{algorithmic}[1]
		\State \textbf{Define} MDP tuple $M = \{S, A, P, R, \gamma\}$
		\State \textbf{Initialize} state space $S$, action space $A$, transition probabilities $P$, reward function $R$, discount factor $\gamma$
		\For{each time step $t$}
		\State Observe current state $s_t = \{D_{i,t}, C_{i,t}, T_{i,t} | i = 1, ..., N\}$
		\State Choose action $a_t = \{x_{i,k,t} | i = 1, ..., N; k \in E \cup C\}$
		\State Execute action $a_t$
		\State Observe new state $s_{t+1}$ and reward $r_t = -(\alpha_1 T_t + \alpha_2 E_t)$
		\State Update MDP model based on observed transition and reward
		\EndFor
	\end{algorithmic}
\end{algorithm}

\subsection{AC-DQN}

AC-DQN algorithm extends traditional deep Q-learning by incorporating an action constraint mechanism, which is crucial for enforcing system limitations in complex serverless environments. This mechanism is implemented through an action constraint function $f_{constraint}: A \rightarrow \mathbb{R}$, which maps actions to penalty values:
\begin{equation}
	f_{constraint}(a_j) = 
	\begin{cases}
		-\lambda, & \text{if } T_{total}(a_j) > T_{max}, \\
		0, & \text{otherwise},
	\end{cases}
\end{equation}
where $\lambda$ is a large positive constant (e.g., 1000), $T_{total}(a_j)$ computes the total delay for action $a_j$, and $T_{max}$ is the maximum allowable delay. This function effectively creates a discontinuity in the action-value space, steering the learning process away from infeasible actions.

The Q-network $Q(s,a;\theta)$ is parameterized by $\theta$ and architecturally consists of an input layer $\mathbb{R}^d \rightarrow \mathbb{R}^{n_1}$, where $d$ is the state dimension, followed by two hidden layers $\mathbb{R}^{n_1} \rightarrow \mathbb{R}^{n_2} \rightarrow \mathbb{R}^{n_3}$, and an output layer $\mathbb{R}^{n_3} \rightarrow \mathbb{R}^{|A|}$, where $|A|$ is the cardinality of the action space. The network is optimized by minimizing the loss function:
\begin{equation}
	L(\theta) = \mathbb{E}_{(s,a,r,s') \sim \mathcal{D}}[(y - Q(s,a;\theta))^2],
\end{equation}
where $\mathcal{D}$ is the experience replay buffer and $y = r + \gamma \max_{a'} (Q(s',a';\theta^-) + f_{constraint}(a'))$ is the target Q-value. Here, $\theta^-$ represents the parameters of a target network, which is periodically updated to stabilize training.

The inclusion of $f_{constraint}(a')$ in the target Q-value calculation is a key innovation of AC-DQN. It allows the constraint information to be propagated through the temporal difference learning process, effectively shaping the Q-function landscape to inherently avoid infeasible actions. This approach contrasts with methods that apply constraints only at the action selection stage, as it embeds the constraints into the learned value function itself.

Algorithm~\ref{alg:ac_dqn} outlines the AC-DQN training process. The algorithm interleaves interaction with the environment, storage of experiences, and neural network updates. The action selection process (line 6) incorporates both the learned Q-values and the constraint function, ensuring that the agent respects system limitations even during exploration. The use of a separate target network (lines 10-12) and the periodic update of its parameters (lines 13-17) are standard techniques in deep Q-learning to improve stability.

\begin{algorithm}
	\caption{AC-DQN Training Process}
	\label{alg:ac_dqn}
	\begin{algorithmic}[1]
		\State Initialize $Q(s,a;\theta)$ and $Q(s,a;\theta^-)$ with $\theta, \theta^- \sim \mathcal{N}(0, \sigma^2)$
		\State Initialize replay memory $\mathcal{D} \leftarrow \{\}$ with capacity $N$
		\For{episode $e = 1$ to $M$}
		\State Initialize state $s_1 \in \mathcal{S}$
		\For{$t = 1$ to $T$}
		\If{$\text{Uniform}(0,1) < 1-\epsilon$}
		\State $a_t \leftarrow \arg\max_{a\in\mathcal{A}}(Q(s_t,a;\theta) + f_{const}(a))$
		\Else
		\State $a_t \leftarrow \text{Uniform}(\mathcal{A})$
		\EndIf
		\State Execute $a_t$, observe $r_t \in \mathbb{R}$ and $s_{t+1} \in \mathcal{S}$
		\State $\mathcal{D} \leftarrow \mathcal{D} \cup \{(s_t, a_t, r_t, s_{t+1})\}$
		\State Sample minibatch $\mathcal{B} \sim \text{Uniform}(\mathcal{D})$ where $|\mathcal{B}| = B$
		\For{$(s_j, a_j, r_j, s_{j+1}) \in \mathcal{B}$}
		\If{$s_{j+1}$ is terminal}
		\State $y_j \leftarrow r_j$
		\Else
		\State $y_j \leftarrow r_j + \gamma \max_{a'\in\mathcal{A}}(Q(s_{j+1},a';\theta^-) + f_{const}(a'))$
		\EndIf
		\EndFor
		\State $\theta \leftarrow \theta - \alpha \nabla_\theta \frac{1}{B}\sum_{j=1}^B (y_j - Q(s_j,a_j;\theta))^2$
		\State $s_t \leftarrow s_{t+1}$
		\EndFor
		\If{$e \bmod \tau = 0$}
		\State $\theta^- \leftarrow \theta$
		\EndIf
		\EndFor
	\end{algorithmic}
\end{algorithm}

\subsection{Security Mechanism Integration}
To ensure data integrity and privacy in serverless multi-cloud edge environments, we devise a robust security mechanism that leverages asymmetric cryptography and cryptographic hash functions, optimizing the balance between security and computational overhead. The security protocol, defined as $\Pi = (\text{KeyGen}, \text{Enc}, \text{Dec}, \text{Hash}, \text{Verify})$, includes RSA key generation, encryption, decryption, MD5 hashing, and verification functions. For a task $\tau_{ij}$ offloaded from MEC node $k$ to CC node $k'$, the security process is as follows: (1) $\text{KeyGen}(1^\lambda) \rightarrow (pk_{k'}, sk_{k'})$; (2) $c_{ij} = \text{Enc}(m_{ij}, pk_{k'})$; (3) $h_{ij} = \text{Hash}(m_{ij})$; (4) Transmit $(c_{ij}, h_{ij})$ from node $k$ to $k'$; (5) $m'_{ij} = \text{Dec}(c_{ij}, sk_{k'})$; (6) $h'_{ij} = \text{Hash}(m'_{ij})$; and (7) $v_{ij} = \text{Verify}(h_{ij}, h'_{ij})$. Here, $m_{ij}$ represents the original task data, $c_{ij}$ is the encrypted data, $h_{ij}$ and $h'_{ij}$ are hash values, and $v_{ij}$ is the verification result.

We model the security-related overhead in both time and energy domains. Let $\phi_{ij}^{enc}$, $\phi_{ij}^{dec}$, and $\omega_{ij}$ represent the number of CPU cycles required for encryption, decryption, and integrity verification (hashing and verification) respectively. The security time overhead $T_{ij}^{sec}$ and energy overhead $E_{ij}^{sec}$ for task $\tau_{ij}$ are formulated as:
\begin{equation}
T_{ij}^{sec} = \frac{\phi_{ij}^{enc} + \omega_{ij}}{f_k} + \frac{\phi_{ij}^{dec} + \omega_{ij}}{f_{k'}},
\end{equation}
\begin{equation}
E_{ij}^{sec} = \mu_k(\phi_{ij}^{enc} + \omega_{ij}) + \mu_{k'}(\phi_{ij}^{dec} + \omega_{ij})
\end{equation},
where $f_k$ and $f_{k'}$ are the CPU frequencies of nodes $k$ and $k'$ respectively, and $\mu_k$ and $\mu_{k'}$ are their respective energy consumption coefficients per CPU cycle.

The security overhead is incorporated into the overall system cost function, allowing \sysname to make informed decisions that balance security requirements with performance and energy efficiency:
\begin{equation}
	C_t = \alpha_1(T_t + \sum_{i,j} T_{ij}^{sec}) + \alpha_2(E_t + \sum_{i,j} E_{ij}^{sec}).
\end{equation}

\subsection{Adaptive Exploration Strategy}

In reinforcement learning, the exploration-exploitation dilemma is a fundamental challenge. We address this through an adaptive $\epsilon$-greedy exploration strategy, which dynamically adjusts the exploration rate based on the learning progress.

Let $\pi_\epsilon(a|s)$ denote our $\epsilon$-greedy policy, defined as:
\begin{equation}
	\pi_\epsilon(a|s) = 
	\begin{cases}
		1 - \epsilon + \frac{\epsilon}{|\mathcal{A}|}, & \text{if } a = \operatorname{argmax}_{a'}Q(s,a'), \\
		\frac{\epsilon}{|\mathcal{A}|}, & \text{otherwise},
	\end{cases}
\end{equation}
where $|\mathcal{A}|$ is the cardinality of the action space. The exploration rate $\epsilon$ is adapted over time according to:

\begin{equation}
\epsilon(t) = \max(\epsilon_{\text{min}}, \epsilon_0 \cdot \epsilon_{\text{decay}}^t),
\end{equation}
where $\epsilon_0$ is the initial exploration rate, $\epsilon_{\text{decay}} \in (0,1)$ is the decay factor, and $\epsilon_{\text{min}}$ is the minimum exploration rate. This formulation ensures a gradual transition from exploration to exploitation while maintaining a baseline level of exploration throughout the learning process.

\subsection{Performance Optimization Techniques}

To enhance the stability and efficiency of learning in complex serverless multi-cloud edge environments, we incorporate several optimizations.

\textbf{Experience replay:}
We maintain a replay buffer $\mathcal{D} = \{e_1, e_2, ..., e_N\}$, where each experience $e_i = (s_i, a_i, r_i, s'_i)$ is a tuple of state, action, reward, and next state. During training, we sample mini-batches $\mathcal{B} \subset \mathcal{D}$ uniformly: $\mathcal{B} \sim \text{Uniform}(\mathcal{D}, n)$,
where $n$ is the batch size. This technique breaks the temporal correlations in the data, reducing the variance of updates and improving learning stability.

\textbf{Target network:}
We employ a target network $Q(s,a;\theta^-)$ alongside the primary Q-network $Q(s,a;\theta)$. The target network parameters $\theta^-$ are updated periodically using a soft update mechanism:
\begin{equation}
\theta^- \leftarrow \tau \theta + (1 - \tau)\theta^-,
\end{equation}
where $\tau \in (0,1]$ is the update rate. This approach stabilizes the learning targets and mitigates the risk of divergence.

\textbf{Prioritized experience replay:}
We extend the experience replay mechanism by assigning priorities to experiences based on their temporal-difference (TD) error. The probability of sampling an experience $e_i$ is given as $P(i) = \frac{p_i^\alpha}{\sum_1^k p_k^\alpha}$,
where $p_i = |\delta_i| + \epsilon$ is the priority of experience $i$, $\delta_i$ is the TD-error, $\epsilon$ is a small positive constant to ensure non-zero sampling probabilities, and $\alpha \in [0,1]$ determines the degree of prioritization.

\textbf{Dueling network architecture:}
We decompose the Q-function into separate value and advantage streams:
\begin{equation}
	\scriptsize
	Q(s,a;\theta,\alpha,\beta) = V(s;\theta,\beta) + \left(A(s,a;\theta,\alpha) - \frac{1}{|\mathcal{A}|}\sum_{a'\in\mathcal{A}}A(s,a';\theta,\alpha)\right),
\end{equation}
where $V(s;\theta,\beta)$ is the state-value function and $A(s,a;\theta,\alpha)$ is the advantage function. This architecture allows for more efficient learning of state values and reduces the overestimation of action values.

\begin{figure*}[!t]
	\centering
	\subfloat[System cost vs. MEC capacity]{\includegraphics[width = 0.46\linewidth]{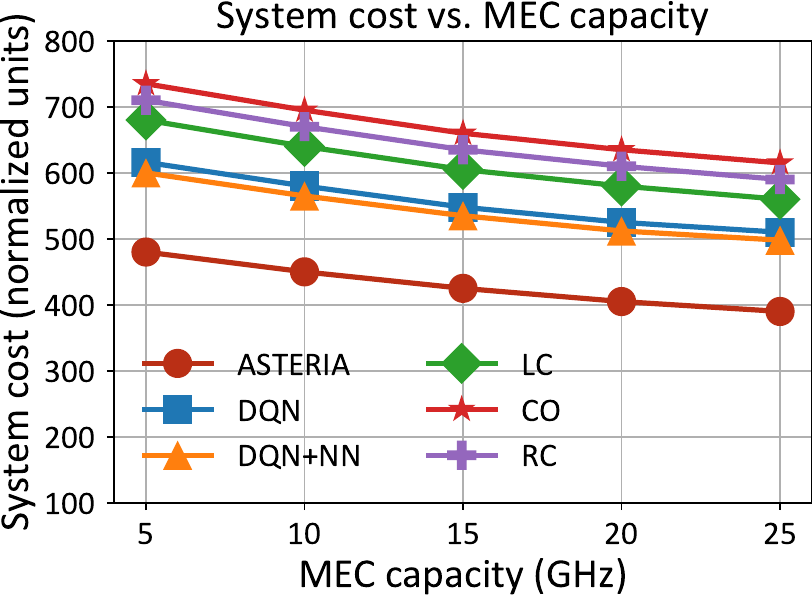}} \hspace{3mm}
	\subfloat[Task completion time vs. MEC capacity]{\includegraphics[width = 0.46\linewidth]{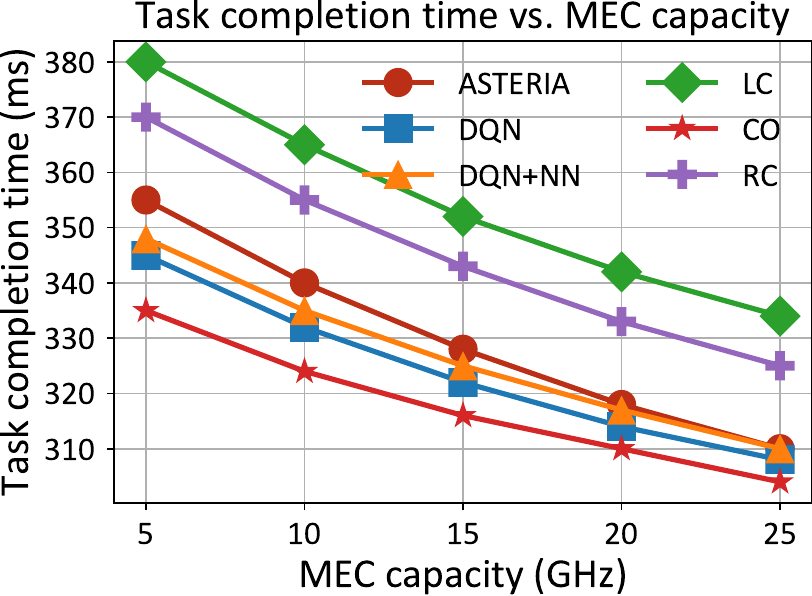}}  \\
	\subfloat[Energy efficiency vs. MEC capacity]{\includegraphics[width = 0.46\linewidth]{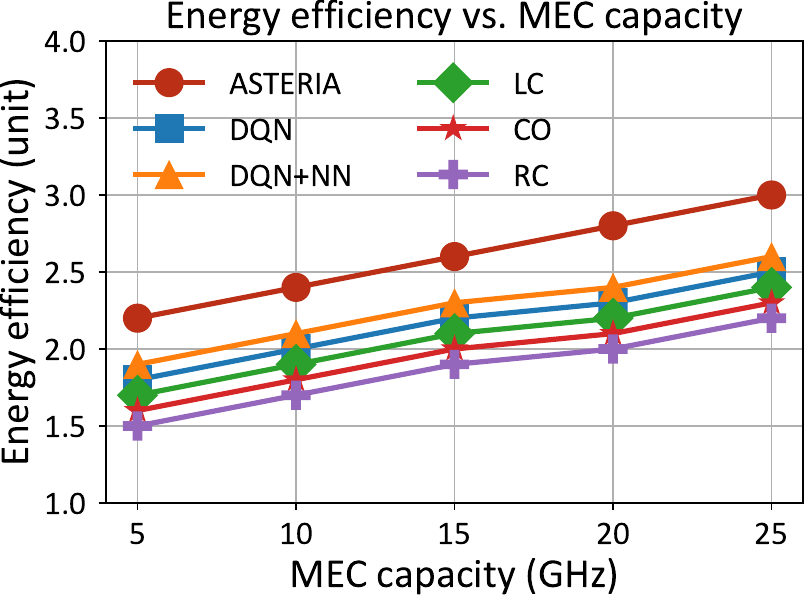}} \hspace{3mm}
	\subfloat[Offloading rate vs. MEC capacity]{\includegraphics[width = 0.46\linewidth]{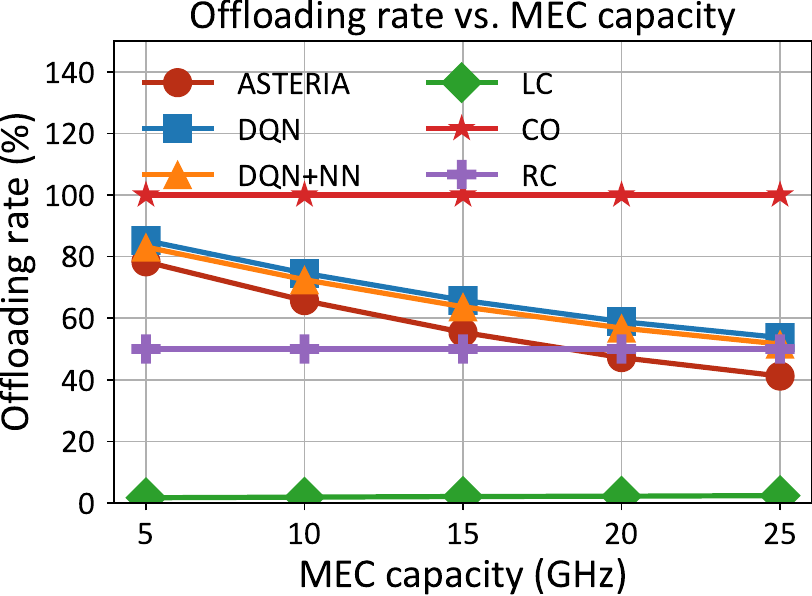}} 
\caption{Performance under different MEC computational capacities, including system cost (a), task completion time (b), energy efficiency (c), and offloading rate (d)} 
	\label{fig:diff_capacity}
\end{figure*}

\section{Experimental Results}
\subsection{Evaluation Setup}
We conducted extensive simulations to evaluate the performance of the proposed \sysname framework, using Python 3.9 for the simulation environment. To provide a comprehensive comparison, we implemented five baseline approaches: MEC Local Computation (LC), where all tasks are processed locally at the MEC; CC Layer Computing Offload (CO), which randomly offloads tasks to the CC layer; Random Computation (RC), which randomly offloads tasks between MEC and CC; a standard Deep Q-Network (DQN) for resource allocation; and an enhanced DQN with Back Propagation Neural Network (DQN+NN). The simulation setup included key parameters such as a system bandwidth of 20 MHz, MEC transmission power of 0.5 W, a channel loss factor of 4, MEC local noise of $10^{-13}$, MEC computation capacity of 10 GHz, and CC computation capacity of 100 GHz. Task data sizes ranged from 1 to 5 GB, with delay requirements between 700 and 800 ms. We assigned equal weights (0.5, 0.5) to delay and energy consumption in our cost function. The network topology included one MEC node and two CC nodes, positioned 1 km and 10 km from the MEC, respectively. To ensure robust results, we ran the simulation for 1000 episodes, with each episode consisting of 100 time slots.

\subsection{Overall Performance}

\begin{figure}[!t]
\centering
\subfloat[System cost vs. number of tasks]{\includegraphics[width = 0.47\linewidth]{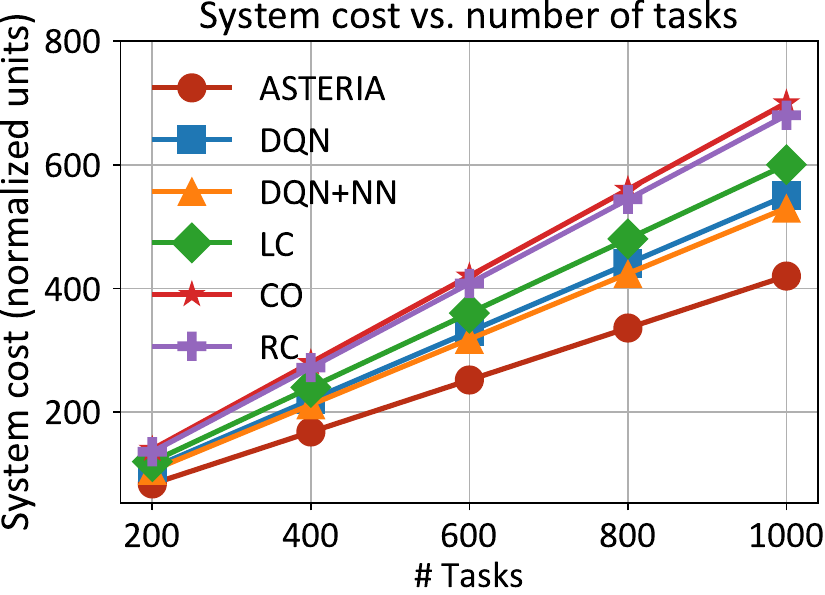}} 
\subfloat[Average delay vs. number of tasks]{\includegraphics[width = 0.47\linewidth]{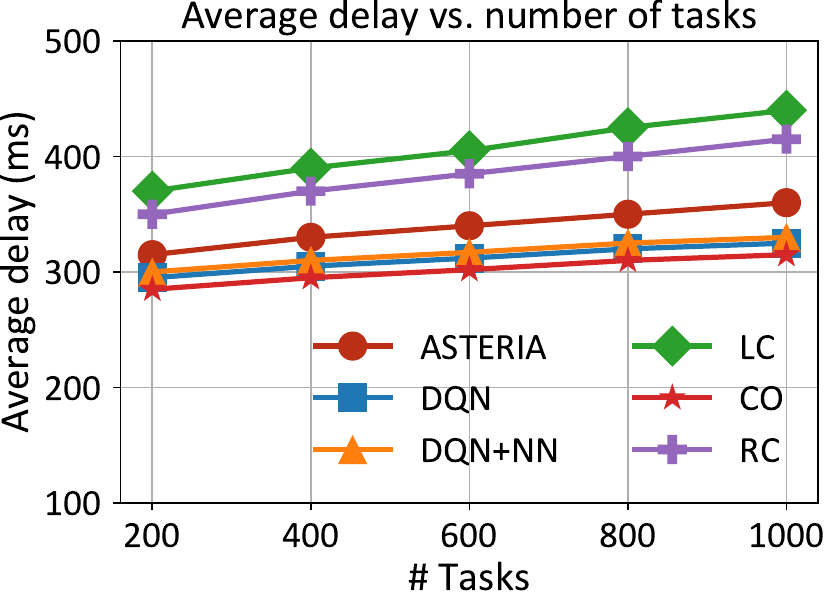}} 
\caption{Performance under different number of tasks, including system cost (a) and average delay (b)}
\label{fig:diff_tasks} 
\end{figure}

We first evaluate the overall performance of \sysname as the number of computational tasks increases from 200 to 1000.
Figure \ref{fig:diff_tasks}(a) illustrates the system cost as the number of tasks increases. \sysname consistently achieves the lowest system cost across all task volumes. At 1000 tasks, \sysname reduces the system cost by 23.6\% compared to DQN, 20.8\% compared to DQN+NN, 30.0\% compared to LC, 40.0\% compared to CO, and 38.2\% compared to RC. This significant improvement demonstrates \sysname's ability to make efficient allocation decisions that balance both delay and energy consumption, even as the system load increases.
As shown in Figure \ref{fig:diff_tasks}(b), it maintains a competitive average delay performance, despite not always achieving the lowest delay. At 1000 tasks, \sysname's average delay is only 14.3\% higher than the best-performing CO method. This slight increase in delay is a trade-off for significant improvements in energy efficiency and overall system cost.

\subsection{Impact of Average Data Size}
\begin{figure}[!t]
\centering
\subfloat[System cost vs. average data size]{\includegraphics[width = 0.47\linewidth]{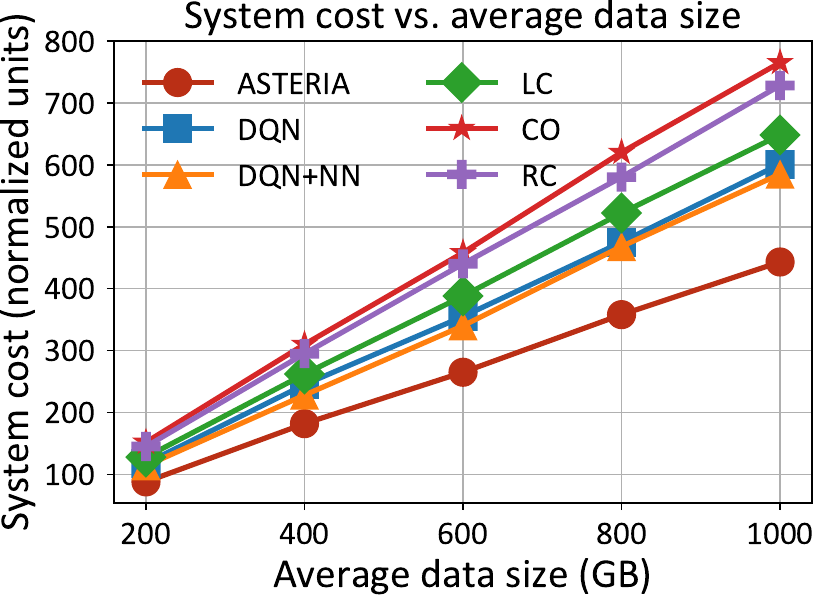}} 
\subfloat[Energy consumption vs. data size]{\includegraphics[width = 0.47\linewidth]{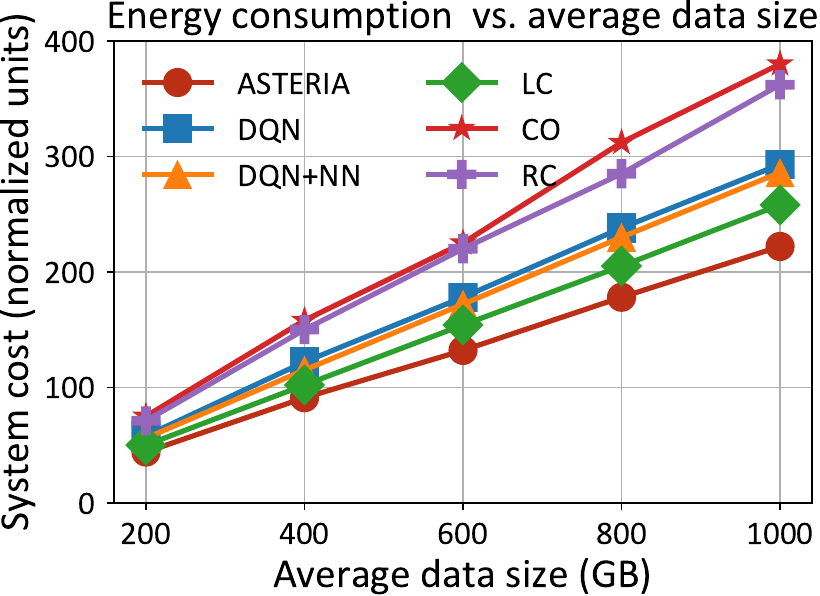}} 
\caption{Performance under different data sizes, including system cost (a) and energy consumption (b)} 
\label{fig:diff_sizes}
\end{figure}

We evaluate the impact of varying average task data sizes, ranging from 1 GB to 5 GB. As shown in Figure \ref{fig:diff_sizes}(a), \sysname consistently achieves the lowest system cost across all data sizes. The performance gap widens as data size increases, with \sysname demonstrating approximately 26.3\% lower system cost compared to DQN at the 5 GB data size, highlighting its ability to efficiently manage larger tasks through intelligent offloading decisions and security-aware resource allocation. Interestingly, the DQN+NN approach shows improved performance for medium-sized data (2-3 GB), slightly narrowing the gap with \sysname; however, this advantage diminishes as data sizes increase.

\subsection{Energy Consumption Analysis}
We further analyzed the energy consumption of different approaches as data size increased, as shown in Figure~\ref{fig:diff_sizes}(b). \sysname consistently demonstrated the lowest energy consumption across all approaches, except for LC, which avoids communication overhead but incurs higher system costs. At 5 GB, \sysname consumed 24.1\% less energy than DQN, 22.3\% less than DQN+NN, 41.5\% less than CO, and 38.7\% less than RC. The energy consumption of CO and RC exhibited greater variability, likely due to their random offloading decisions, particularly as data size increased. This underscores the importance of intelligent resource allocation, where \sysname excels by effectively balancing energy efficiency with overall system performance, especially for larger tasks.

\subsection{Impact of MEC Computational Resources}
To evaluate \sysname's adaptability and efficiency under varying edge computing capabilities, we conducted experiments by adjusting the MEC computational capacity from 5 GHz to 25 GHz. We maintained a constant workload of 1000 tasks, each with an average data size of 3 GB, across all MEC capacities. The  measured metrics included system cost, task completion time, energy efficiency, and offloading rate.

As illustrated in Figure~\ref{fig:diff_sizes}(a), \sysname consistently outperforms other methods across all MEC capacities. It achieves up to 34.7\% lower system costs at lower capacities, with a still significant advantage at higher capacities. Task completion times are competitive, with \sysname being only 5.8\% slower than the best-performing method at 5 GHz, narrowing to 2.1\% at 25 GHz. In terms of energy efficiency, \sysname completes up to 27.3\% more tasks per unit of energy at the highest MEC capacity. Its adaptive offloading strategy is particularly effective, offloading more tasks to the cloud at lower MEC capacities (78.3\% at 5 GHz) and decreasing to 41.2\% at 25 GHz. This demonstrates \sysname's ability to make context-aware decisions, utilizing available edge resources while balancing performance and energy efficiency. These results highlight \sysname's robustness and scalability across diverse edge computing scenarios, making it well-suited for dynamic serverless multi-cloud environments where computational resources vary across nodes.

\section{Conclusion}
This paper has introduced \sysname, a framework for secure resource allocation in serverless multi-cloud edge environments, leveraging action-constrained deep reinforcement learning. \sysname effectively balances task offloading, resource allocation, and security requirements through its innovative algorithm, adaptive security mechanisms, and performance optimization techniques. Extensive simulations demonstrated its superior performance, consistently outperforming existing methods in system cost, energy efficiency, and adaptability across various scenarios. As edge computing evolves, it offers a promising direction for future research in serverless multi-cloud edge computing, potentially enabling more efficient and secure distributed systems at scale~\cite{wu2022echohand,duan2025rethinking,wu2021toward,xu2024sok,wu2020liveness}. As a potential future direction, we are looking forward to extending our \sysname to improve the performance of various applications such as large language models~\cite{lin2023pushing,wu2024semantic}, distributed learning system~\cite{lin2024adaptsfl,zhang2025lcfed,lin2024split}, software engineering, and security~\cite{2024-LLM4CodeSum,2024-Survey-of-Source-Code-Search,2024-EACS,2024-ESALE,2024-TranCS,2023-AST4PL,2019-MAF,2023-BADCODE,2024-EliBadCode,2024-MIMIC,2024-Distillation-in-Mitigating-Backdoors-in-Pre-trained-Encoder,2022-RULER}.

\section*{Acknowledgments}
This work is supported by the Natural Science Foundation of Sichuan Province, China (Grant No. 2023NSFSC1411), and the Sichuan Science and Technology Program, China (Grant No. 2023ZYD0145). Also, this research is supported by the National Research Foundation, Singapore, and Infocomm Media Development Authority under its Future Communications Research \& Development Programme, Defence Science Organisation (DSO) National Laboratories under the AI Singapore Programme (FCP-NTU-RG-2022-010 and FCP-ASTAR-TG-2022-003), Singapore Ministry of Education (MOE) Tier 1 (RG87/22), the NTU Centre for Computational Technologies in Finance (NTU-CCTF), and Seitee Pte Ltd. Jianfei Sun and Qiang Gao contribute equally to this work. 
\bibliographystyle{IEEEtran}
\bibliography{lib.bib}

\end{document}